\documentclass[10pt,twocolumn]{article}
\pdfoutput=1
\usepackage{times}
\usepackage{epsfig}
\usepackage{graphicx}
\usepackage{amsmath}
\usepackage{amssymb}
\usepackage{xcolor}
\usepackage{algorithm}
\usepackage{algpseudocode}
\usepackage[font=small,labelfont=bf,tableposition=bottom,figureposition=bottom]{caption}

\DeclareMathOperator*{\argmin}{arg\,min}

\begin{document}

\title{Computing the Testing Error without a Testing Set}

\author{Ciprian A. Corneanu\\
Univ. Barcelona, Tawny\\
\and
Meysam Madadi\\
CVC, UAB\\
\and
Sergio Escalera \\
Univ. Barcelona, CVC\\
\and
Aleix M. Martinez\\
OSU, Amazon\\
}
\date{}
\maketitle
\setcounter{footnote}{1}


\begin{abstract}
Deep Neural Networks (DNNs) have revolutionized computer vision. We now have DNNs that achieve top (performance) results in many problems, including object recognition, facial expression analysis, and semantic segmentation, to name but a few. The design of the DNNs that achieve top results is, however, non-trivial and mostly done by trail-and-error. That is, typically, researchers will derive many DNN architectures (i.e., topologies) and then test them on multiple datasets. However, there are no guarantees that the selected DNN will perform well in the real world. One can use a testing set to estimate the performance gap between the training and testing sets, but avoiding overfitting-to-the-testing-data is almost impossible. Using a sequestered testing dataset may address this problem, but this requires a constant update of the dataset, a very expensive venture. Here, we derive an algorithm to estimate the performance gap between training and testing that does not require any testing dataset. Specifically, we derive a number of persistent topology measures that identify when a DNN is learning to generalize to unseen samples. This allows us to compute the DNN's testing error on unseen samples, even when we do not have access to them. We provide extensive experimental validation on multiple networks and datasets to demonstrate the feasibility of the proposed approach.
\end{abstract}

\section{Introduction}

\footnotetext{Let $\rho_{train}$ and $\rho_{test}$ be the performance of an algorithm computed using a training and testing set, respectively; $\hat{\rho}_{test}$ is the estimated testing error computed without any testing data. The performance metric may be classification accuracy, F1-score, Intersection-over-Union (IoU), etc.}

Deep Neural Networks (DNNs) are algorithms capable of identifying complex, non-linear mappings, $f(.)$, between an input variable ${\bf x}$ and and output variable ${\bf y}$, i.e., $f({\bf x})={\bf y}$ \cite{lecun2015deep}. Each DNN is defined by its unique topology and loss function. Some well-known models are \cite{simonyan2014very, he2016deep, krizhevsky2012imagenet}, to name but a few.

Given a well-curated dataset with $n$ samples, $\mathcal{X}={ \{\bf x}_i,{\bf y}_i \}_{i=1}^{n}$, we can use DNNs to find an estimate of the functional mapping $f({\bf x}_i)={\bf y}_i$. Let us refer to the estimated mapping function as $\hat{f}(.)$. Distinct estimates, $\hat{f}_j(.)$, will be obtained when using different DNNs and datasets. Example datasets we can use to this end are \cite{deng2009imagenet, fabian2016emotionet, everingham2015pascal}, among many others.

Using datasets such as these to train DNNs has been very fruitful. DNNs have achieved considerable improvements in a myriad of, until recently, very challenging tasks, e.g., \cite{krizhevsky2012imagenet,vinyals2015show}.

Unfortunately, we do not generally know how the estimated mapping functions $\hat{f}_j(.)$ will perform in the real world, when using independent, unseen images.

\begin{figure*}
    \centering
    \includegraphics[width=.99\linewidth]{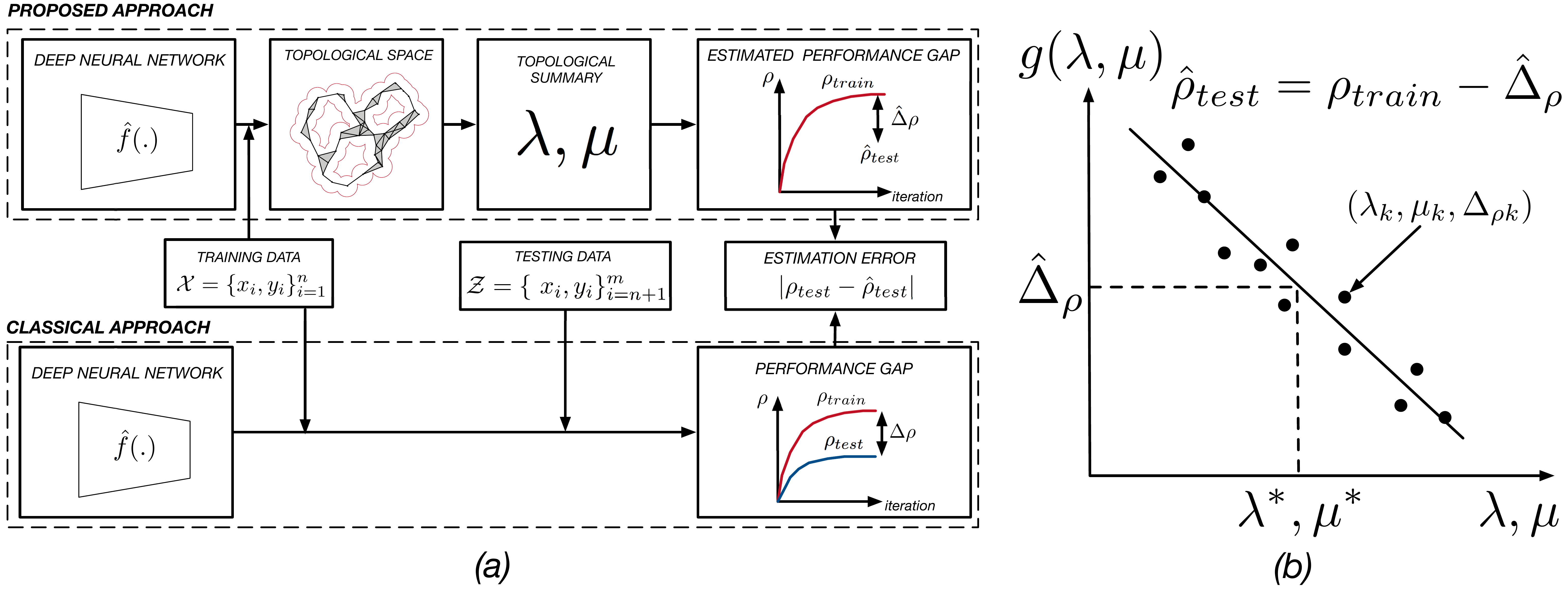}
    \caption{(a) We compute the test performance of any Deep Neural Network (DNN) on any computer vision problem using no testing samples (top); neither labelled nor unlabelled samples are necessary. This in sharp contrast to the classical computer vision approach, where model performance is calculated using a curated test dataset (bottom). (b) The persistent algebraic topological summary ($\lambda^{*}$, $\mu^{*}$) given by our algorithm ($x$-axis) against the performance gap $\Delta\rho$ between training and testing performance ($y$-axis).}
    \label{fig:overview}
\end{figure*}

The classical way to address this problem is to use a testing dataset, Figure \ref{fig:overview}(a, bottom). The problem with this approach is that, in many instances, the testing set is visible to us, and, hence, we keep modifying the DNN topology until it works on this testing dataset. This means that we overfit to the testing data and, generally, our algorithm may not be the best for truly unseen samples.

To resolve this issue, we can use a sequestered dataset. This means that a third-party has a testing dataset we have never seen and we are only able to know how well we perform on that dataset once every several months. While this does tell us how well our algorithm performs on previously unseen samples, we can only get this estimate sporadically. And, importantly, we need to rely on someone else maintaining and updating this sequestered testing set. Many such sequestered datasets do not last long, because maintaining and updating them is a very costly endeavour.

In the present paper, we introduce an approach that resolves these problems. Specifically, we derive an algorithm that gives an accurate estimate of the performance gap between our training and testing error, {\em without the need of any testing dataset}, Figure \ref{fig:overview}(a, top). That means we do not need to have access to any labelled or unlabelled data. Rather, our algorithm will give you an accurate estimate of the performance of your DNN on independent, unseen sample.

Our key idea is to derive a set of topological summaries measuring persistent topological maps of the behavior of DNNs across computer vision problems. Persistent topology has been shown to correlate with generalization error in classification  \cite{corneanu2019does}, and as a method to theoretically study and explain DNNs' behavior \cite{bergomi2019towards,corneanu2019does,zador2019critique}. The hypothesis we are advancing is that the generalization gap is a function of the inner-workings of the network, here represented by its functional topology and described through topological summaries. We propose to regress this function and use it to estimate test performance based only on training data.

Figure \ref{fig:overview}(b) shows an example. In this plot, the $x$-axis shows a linear combination of persistent topology measures of DNNs. The $y$-axis in this plot is the value of the performance gap when using these DNNs on multiple computer vision problems. As can be seen in this figure, there is a linear relationship between our proposed topological summaries and the DNN's performance gap. This means that knowing the value of our topological summaries is as good as knowing the performance of the DNN on a sequestered dataset, but without any of the drawbacks mentioned above -- no need to depend on an independent group to collect, curate, and update a testing set.

We start with a set of derivations of the persistent topology measures we perform on DNNs (Section \ref{Section: Topological summaries}), before using this to derive our algorithm (Section \ref{Section: Algorithm}). We provide a discussion of related work (Section \ref{sec:related_work}) and extensive experimental evaluations on a variety of DNNs and computer vision problems, including object recognition, facial expression analysis, and semantic segmentation (Sections \ref{sec:experimental_settings} and \ref{sec:results}).

\section{Topological Summaries}\label{Section: Topological summaries}

A DNN is characterized by its \emph{structure} (i.e., the way its computational graph is defined and trained), and its \emph{function} (i.e, the actual values its components take in response to specific inputs). We focus here on the latter.

To do this, we define DNNs on a topological space. A set of compact descriptors of this space, called \emph{topological summaries}, are then calculated. They measure important properties of the network's behaviour. For example, a summary of the \emph{functional topology} of a network can be used to detect overfitting and perform early-stopping \cite{corneanu2019does}.

Let $A$ be a set. An \emph{abstract simplicial complex} $S$ is a collection of vertices denoted $V(A)$, and a collection of subsets of $V(A)$ called \emph{simplices} that is closed under the subset operation, i.e., if $\sigma \in \psi$ and $\psi \in A$, then $\sigma \in A$.

The dimension of a simplex $\sigma$ is $|\sigma|-1$, where $|\cdot|$ denotes cardinality. A simplex of dimension $n$ is called a $n$-simplex. A $0$-simplex is realized by a single vertex, a $1$-simplex by a line segment (i.e., an edge) connecting two vertices, a $2$-simplex is the filled triangle that connects three vertices, etc.

Let $M=(A, \nu)$ be a metric space -- the association of the set $A$ with a metric $\nu$. Given a distance $\epsilon$, the Vietoris-Rips complex \cite{vietoris1927hoheren} is an abstract simplicial complex that contains all the simplices formed by all pairs of elements $a_i,a_j \in A$ with
\begin{equation}
    \nu(a_i, a_j) < \epsilon,
\end{equation}
for some small $\epsilon>0$, and $i\neq j$.

\begin{figure}
    \centering
    \includegraphics[width=\linewidth]{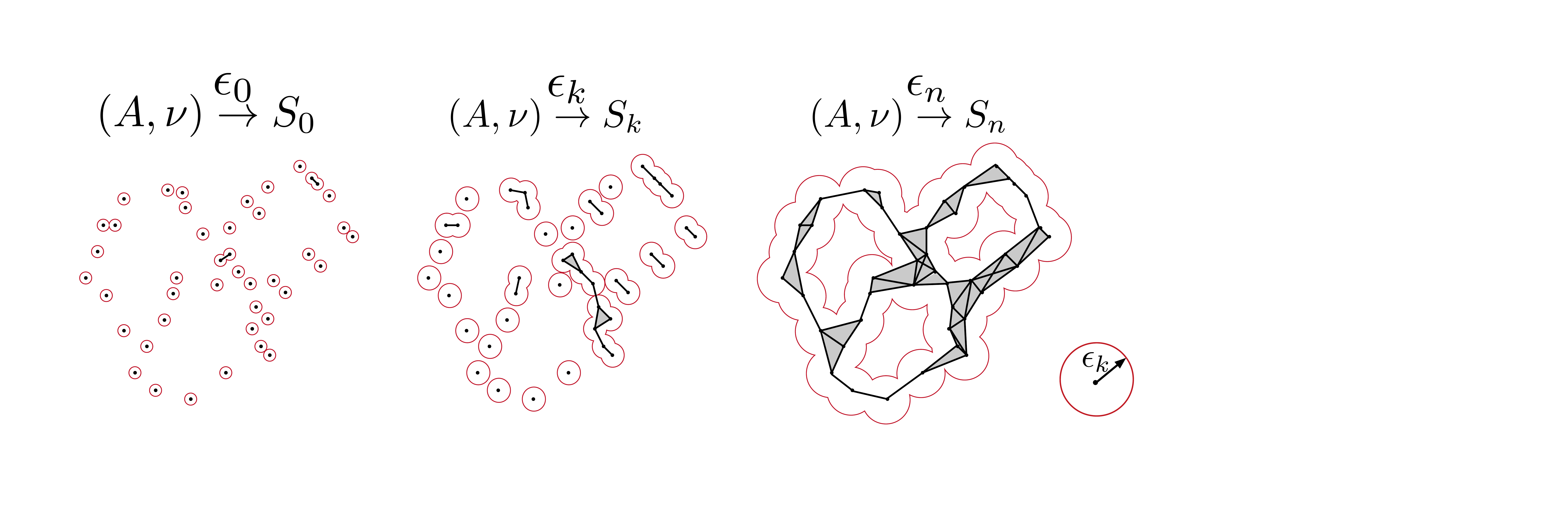}
    \caption{Given a metric space, the Vietoris-Rips filtration creates a nested sequence of simplicial complexes by connecting points situated closer than a predefined distance $\epsilon$. Varying $\epsilon$, we compute persistent properties (cavities) in these simplices. We define a DNN in one such topological space to compute informative data of its behaviour that correlates with its performance on testing data.}
    \label{fig:vietoris_rips_filtration}
\end{figure}

By considering a range of possible distances, $E = \{\epsilon_0, \epsilon_1, \dots \epsilon_k, \dots \epsilon_n\}$, where $0<\epsilon_0 \le \epsilon_1 \le \dots \le \epsilon_k \le \dots \le \epsilon_n$, a Vietoris-Rips filtration yields a collection of simplicial complexes, $\mathcal{S} = \{S_0, S_1, \dots, S_k, \dots, S_n\}$, at multiple scales, Figure \ref{fig:vietoris_rips_filtration} \cite{hatcher2002algebraic}.

\begin{algorithm*}[t]
\caption{Topological Early Stopping.}
\label{algo:alg1}
\begin{algorithmic}
\State{\textbf{Input}: train dataset $\mathcal{X} = \{x_i, y_i \}_{i=1}^n$; DNN partition $A$.}
\Repeat
    \State $\omega$ $\leftarrow{\argmin_{\omega} L(\hat{f}(x;\omega),y)}$ \Comment{Train DNN and estimate $f$ by optimizing loss $L$ over $\mathcal{X}$.}
    \For {all pairs of nodes $(a_p, a_q) \in A$, $p\neq q$}
        \State $\nu_{pq} \leftarrow{ \sum_{i=1}^{n}\frac{(a_{pi}-\overline{a_p})(a_{qi}-\overline{a_q})}{\varsigma_{a_p} \varsigma_{a_q}}}$ \Comment{Compute correlations.}
    \EndFor
    \State $\mathcal{S} \leftarrow{VR(A, \nu)}$ \Comment{Perform Vietoris-Rips filtration and get set of simplicial complexes $\mathcal{S}$}.
    \State  $P \leftarrow \mathcal{PH}(\mathcal{S})$ \Comment{Compute persistent homology $\mathcal{PH}$ over $\mathcal{S}$ and get persistent diagram $P$.}
    \State $\beta_d \leftarrow \{ \sum_{S_k}  \textbf{1}_{cav}, k=1, \dots, n \} $ \Comment{Compute Betti Curves.}
    \State $\widehat{k}_t=\arg\max_{k} \beta_d (S_k)$
    \State $t \leftarrow t+1$
\Until{$\widehat{k}_t>\widehat{k}_{t-1}$.}
\end{algorithmic}
\end{algorithm*}

We are interested in the {\em persistent topology properties} of these complexes across different scales. For this, we compute the $p^{th}-$ persistent homology groups and the Betti numbers $\beta _{p}$, which gives us the ranks of those groups \cite{hatcher2002algebraic}. This means that the Betti numbers compute the number of \emph{cavities} of a topological object.\footnote{Two objects are topologically equivalent if they have the same number of cavities (holes) at each of their dimensions. For example, a donut and a coffee mug are topologically equivalent, because each has a single 2D cavity, the whole in the donut and in the handle of the mug. On the other hand, a torus (defined as the product of two circles, $S^1\times S^1$) has two holes because it is hollow. Hence, a torus is topologically different to a donut and a coffee mug.}

\begin{figure}
    \centering
    \includegraphics[width=\linewidth]{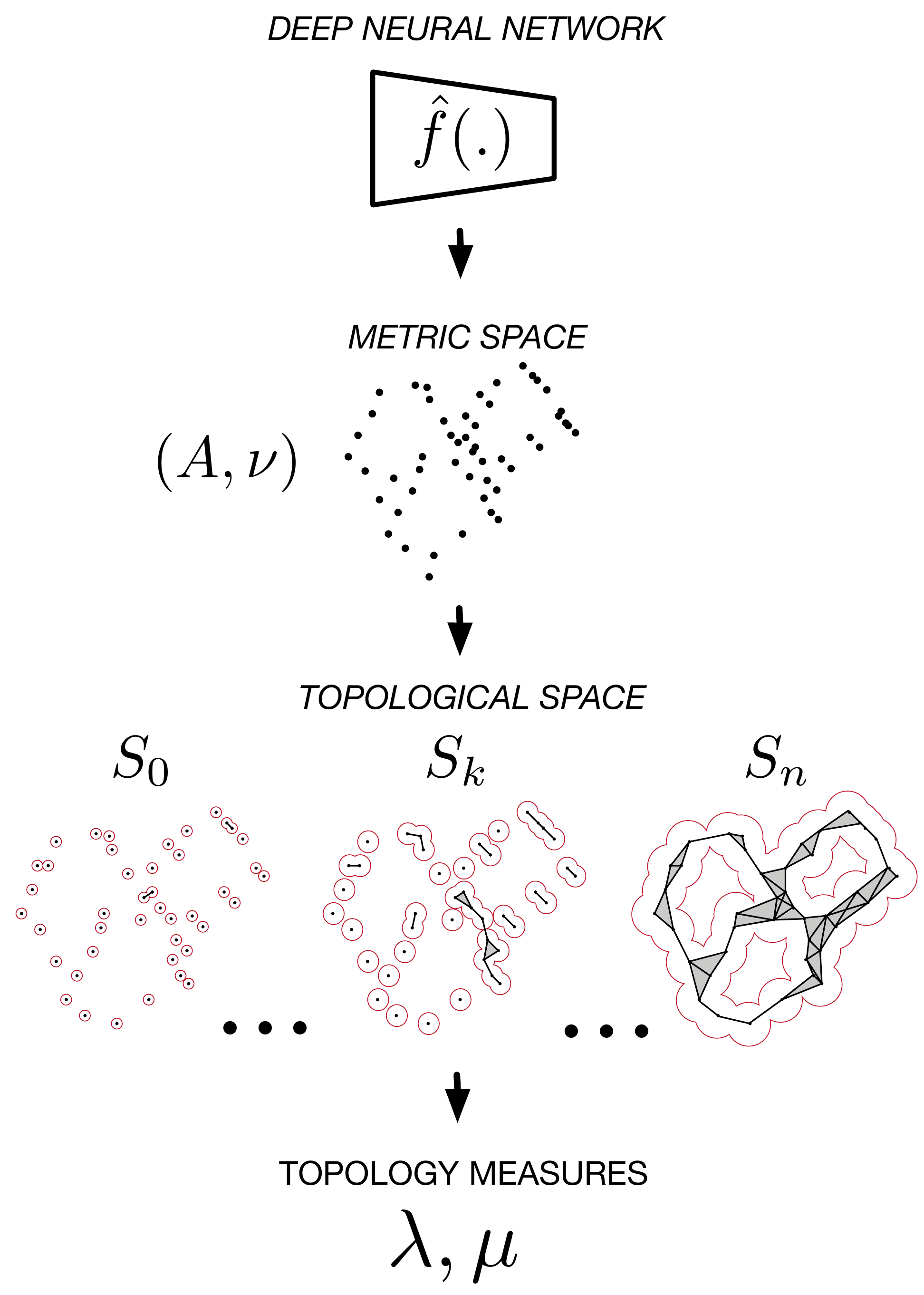}
    \caption{An overview of computing topological summaries from DNNs. We first define a set of nodes $A$ in the network. By computing the correlations between these nodes we project the network into a metric space $(A, \nu)$ from which we obtain a set of simplicial complexes in a topological space through Vietoris-Rips filtration. Persistent homology on this set of simplicial complexes results in a persistence diagram from which topological measures can be computed directly.}
    \label{fig:methodology}
\end{figure}

In DNNs, we can, for example, study how its functional topology varies during training as follows (Fig. \ref{fig:methodology}). First, we compute the correlation of every node in our DNN to every other node at each epoch. Nodes that are highly correlated (i.e., their correlation is above a threshold) are defined as connected, even if there is no actual edge or path connecting them in the network's computational graph. These connections define a simplicial complex, with a number of cavities. These cavities are given by the Betti numbers. We know that the dynamics of low-dimension Betti numbers (i.e., $\beta_1$ and $\beta_2$) is informative over the bias-variance problem (i.e., the generalization vs. memorization problem) \cite{corneanu2019does}. Similarly, it has been shown that these persistence homology measures can be used to study and interpret the data as points in a functional space, making it possible to learn and optimize the estimates $f_j(.)$ defined on the data \cite{bergomi2019towards}.

\section{Algorithm}\label{Section: Algorithm}


Recall $\mathcal{X}=\left\{ {\bf x}_i, {\bf y}_i \right\}_{i=1}^{n}$ is the set of labeled training samples, with $n$ the number of samples. 

Let $a_i$ be the activation value of a particular node in our DNN for a particular input ${\bf x}_i$. 
Passing the sample vectors ${\bf x}_i$ through the network ($i=1,\dots,n$), allows us to compute the correlation between the activation $(a_p, a_q)$ of each pair of nodes which defines the metric $\nu$ of our Vietoris-Rips complex. Formally,
\begin{equation}\label{eq:metric}
\nu_{pq} = \sum_{i=1}^{n} \frac{(a_{pi}-\overline{a_p})(a_{qi}-\overline{a_q})}{\varsigma_{a_p} \varsigma_{a_q}},
\end{equation}
where $\overline{a}$ and $\varsigma_a$ indicate the mean and standard deviation over $\mathcal{X}$.

We represent the results of our persistent homology using a  \emph{persistence diagram}. In our persistence diagram,  each point has as coordinates a set of pairs of real positive numbers $P = \{(\epsilon^i_d,\epsilon^i_b)| (\epsilon^i_d,\epsilon^i_b) \in \mathbb{R}\times\mathbb{R} , i = 1, \dots C\}$, where the subscripts $b$ and $d$ in $\epsilon$ indicate the birth and death distances of a cavity $i$ in the Vietoris-Rips filtration, and $C$ is the total number of cavities.

A filtration of a metric space $(A, \nu)$ is a nested subsequence of complexes that abide by the following rule: $S_0 \subseteq S_1 \subseteq \dots \subseteq S_n = \mathcal{S}$ \cite{zomorodian2005computing}. Thus, this filtration is in fact what defines the persistence diagram of a $k$-dimensional homology group. This is done by computing the creation and deletion of $k$-dimensional homology features. This, in turn, allows us to compute the lifespan homological feature \cite{chazal2009gromov}.

Based on this persistence diagram, we define the \emph{life} of a cavity as the average time (i.e., persistence) in this diagram. Formally,
\begin{equation}
    \lambda = \frac{1}{C}\sum_{i=1}^{C} (\epsilon^i_d-\epsilon^i_b).
    \label{eq:life}
\end{equation}

Similarly, we define its \emph{midlife} as the average density of its persistence. Formally,
\begin{equation}
    \mu = \frac{1}{C}\sum_{i=1}^{C} \frac{\epsilon^i_d+\epsilon^i_b}{2}.
    \label{eq:midlife}
\end{equation}

Finally, we define the {\em linear} functional mapping from these topological summaries to the gap between the training and testing error as,
\begin{equation}
    g\left( \lambda,\mu ; {\bf c} \right) = \widehat{\Delta}{\rho},
    \label{eq:g}
\end{equation}
where $\widehat{\Delta}{\rho}$ is our estimate of the gap between the training and testing errors, and $g\left( \lambda,\mu; {\bf c} \right) = c_1 \lambda + c_2 \mu + c_3$, with $c_i \in \mathbb{R}^+$, and ${\bf c}=(c_1,c_2,c_3)^T$, Figure \ref{fig:overview}(b).

With the above result we can estimate the testing error without the need of any testing data as,
\begin{equation}
    \hat{\rho}_{test} = \rho_{train} - \widehat{\Delta}{\rho},
\end{equation}
where $\rho_{train}$ is the training error computed during training with $\mathcal{X}$.

Given an actual testing dataset $\mathcal{Z}=\{ {\bf x}_i,{\bf y}_i \}_{n+1}^{m}$, we can compute the accuracy of our estimated testing error as,
\begin{equation}\label{Eq:Error}
    Error = |\rho_{test} - \hat{\rho}_{test}|,
\end{equation}
where $\rho_{test}$ is the testing error computed on $\mathcal{Z}$.

The pseudo-code of our proposed approach is shown in Alg. \ref{algo:alg1}.\footnote{Code available at https://github.com/cipriancorneanu/dnn-topology.}

\subsection{Computational Complexity}
Let the binomial coefficient $p=\binom{N+1}{n+1}$ be the number of $n$-simplices of a simplicial complex $S$ (as, for example, would be generated during the Vietoris-Rips filtration illustrated in Fig. \ref{fig:vietoris_rips_filtration}). In order to compute persistent homology of order $n$ on S, one has to compute $rank(\partial_{n+1})$, with $\partial_{n+1} \in \mathbb{R}^{p \times q}$, $p$ the number of $n$-simplices, and $q$ the number of $(n+1)$-simplices. This has polynomial complexity $O(q^a)$, $a>1$.

Fortunately, in Alg. \ref{algo:alg1}, we only need to compute persistent homology of the first order. Additionally, the simplicial complexes generated by the Vietoris-Rips filtration are generally extremely sparse. This means that for typical DNNs, the number of $n$-simplices is way lower than the binomial coefficient defined above. In practice, we have found 10,000 to be a reasonable upper bound for the cardinality of $A$. This is because we define nodes by taking into account structural constraints on the topology of DNNs. Specifically, a node $a_i$ is a random variable with value equal to the mean output of the filter in its corresponding convolutional layer. Having random variables allows us to define correlations and metric spaces in Alg. \ref{algo:alg1}. Empirically, we have found that defining nodes in this way is robust, and similar characteristics, e.g. high correlation, can be found even if a subset of filters is randomly selected. For smaller, toy networks there is previous evidence \cite{corneanu2019does} that supports that functional topology defined in this way is informative for determining overfitting in DNNs.

Finally, the time it takes to compute persistent homology, and consequently, the topological summaries, $\lambda$ and $\mu$, is 5 minutes and 15 seconds for VGG16, one of the most extended networks in our analysis. This corresponds to a single iteration of Alg. \ref{algo:alg1} (the for-loop that iterates over $k$), excluding training, on a single 2.2 GHz Intel Xeon CPU.

\section{Related Work}
\label{sec:related_work}

Topology measures have been previously used to identify over-fitting in DNNs. For example, using lower dimensional Betti curves (which calculates the cavities) of the functional (binary) graph of a network \cite{corneanu2019does}, which can be used to perform early stopping in training and detect adversarial attacks. Other topological measures, this time for characterizing and monitoring {\em structural} properties, have been used for the same purpose \cite{rieck2018neural}.

Other works tried to address the crucial question of how the generalization gap can be predicted from training data and network parameters \cite{barrett2019emotional, arora2018stronger, neyshabur2017exploring, jiang2018predicting}. For example, a metric based on the ratio of the margin distribution at the output layer of the network and a spectral complexity measure related to the network’s Lipschitz constant has been proposed \cite{bartlett2017spectrally}. In \cite{neyshabur2017exploring}, the authors developed bounds on the generalization gap based on the product of norms of the weights across layers. In \cite{arora2018stronger}, the authors developed bounds based on noise stability properties of networks showing that more stability implies better generalization. And, in \cite{jiang2018predicting}, the authors used the notion of margin in support vector machines to show that the normalized margin distribution across a DNN's layers is a predictor of the generalization gap.

\section{Experimental Settings}\label{Sec: Settings}
\label{sec:experimental_settings}

\begin{figure}
    \centering
    \includegraphics[width=\linewidth]{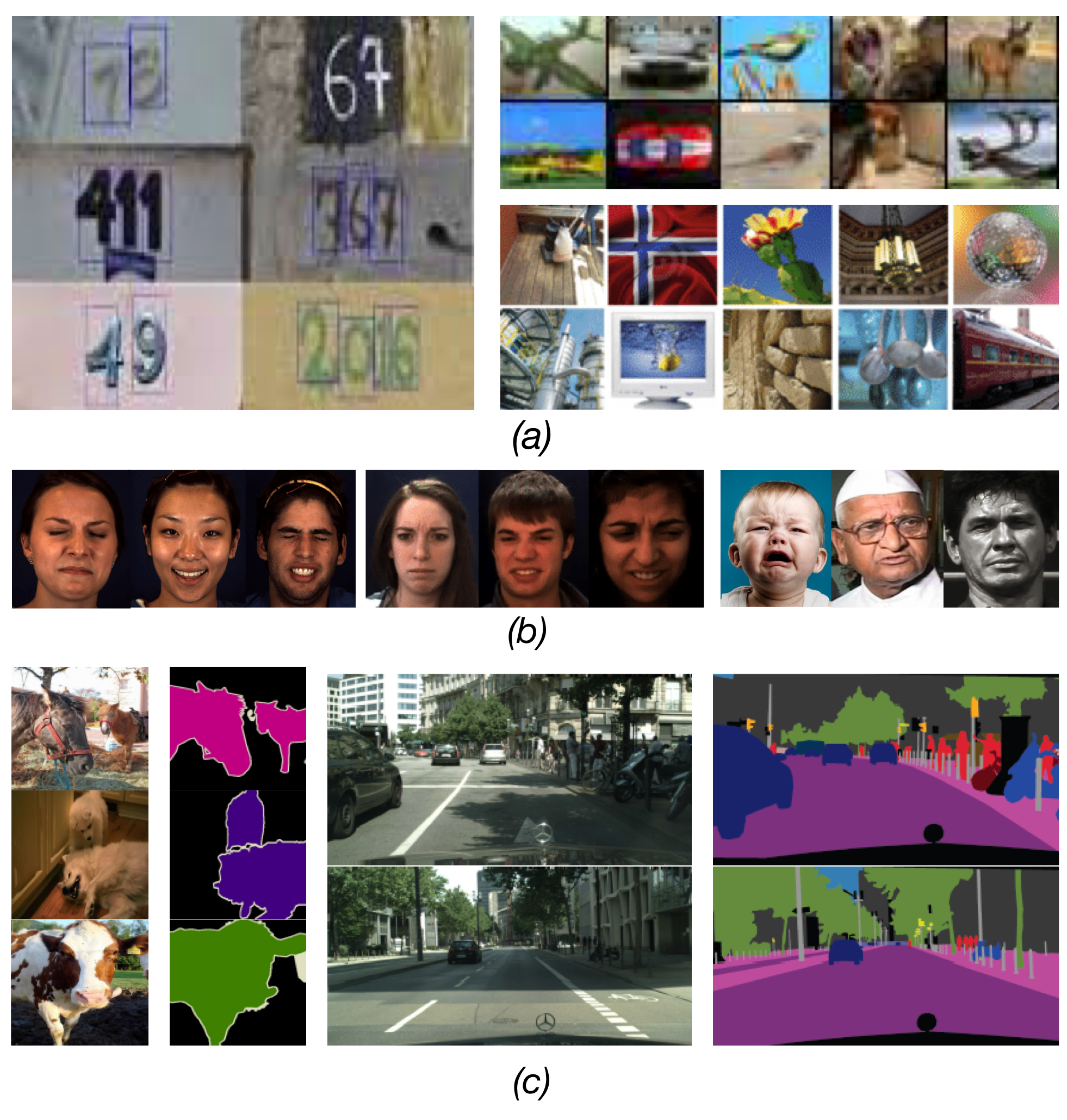}
    \caption{We evaluate the proposed method on three different vision problems. (a) Object recognition is a standard classification problem consisting in categorizing objects. We evaluate on datasets of increasing difficulty, starting with real world digit recognition and continuing towards increasingly challenging category recognition. (b) AU recognition involves recognizing local, sometimes subtle patterns of facial muscular articulations. Several AUs can be present at the same time, making it a multi-label classification problem. (c) In semantic segmentation, one has to output a dense pixel categorization that properly captures complex semantic structure of an image.}
    \label{fig:data}
\end{figure}

We have derived an algorithm to compute the testing accuracy of a DNN that does not require access to any testing dataset. This section provides extensive validation of this algorithm. We apply our algorithm in three fundamental problems in computer vision: object recognition, facial action unit recognition, and semantic segmentation, Figure \ref{fig:data}.

\subsection{Object Recognition}

Object recognition is one of the most fundamental and studied problems in computer vision. Many large scale databases exist, allowing us to provide multiple evaluations of the proposed approach.

To give an extensive evaluation of our algorithm, we use four datasets: CIFAR10 \cite{krizhevsky2009learning}, CIFAR100 \cite{krizhevsky2009learning}, Street View House Numbers (SVHN) \cite{netzer2011reading}, and ImageNet \cite{deng2009imagenet}. In the case of ImageNet, we have used a subset with roughly $120,000$ images split in 200 object classes.

We evaluate the performance of several DNNs by computing the classification accuracy, namely the number of predictions the model got right divided by the total number of predictions, Figure \ref{fig:results_object_recognition} and Tables \ref{tab:eval_or} \& \ref{tab:obj_rec_ldo}.

\subsection{Facial Action Unit Recognition}

Facial Action Unit (AU) recognition is one of the most difficult tasks for DNNs, with humans significantly over-performing even the best algorithms \cite{barrett2019emotional,benitez2017emotionet}.

Here, we use BP4D \cite{zhang2014bp4d}, DISFA \cite{mavadati2013disfa}, and EmotionNet \cite{fabian2016emotionet}, of which, in this paper, we use a subset of 100,000 images. And, since this is a binary classification problem, we are most interested in computing precision and recall, Figure \ref{fig:results_au_recognition} and Tables \ref{tab:eval_aur} \& \ref{tab:au_rec_ldo}.

\subsection{Semantic Segmentation}

Semantic segmentation is another challenging problem in computer vision. We use Pascal-VOC \cite{pascal-voc-2012} and Cityscapes \cite{cordts2016Cityscapes}. The version of Pascal-VOC used consists of 2,913 images, with pixel based annotations for 20 classes. The Cityscapes dataset focuses on semantic understanding of urban street scenes \cite{cordts2016Cityscapes}. It provides 5,000 images with dense pixel annotations for 30 classes.

Semantic segmentation is evaluated using union-over-intersection (IoU\footnote{Also known as the Jaccard Index., which counts the number of pixels common between the ground truth and prediction segmentation masks divided by the total number of pixels present across both masks.}), Figure \ref{fig:results_ss} and Table \ref{tab:eval_ss}.

\begin{figure*}
    \centering
    \includegraphics[width=\linewidth]{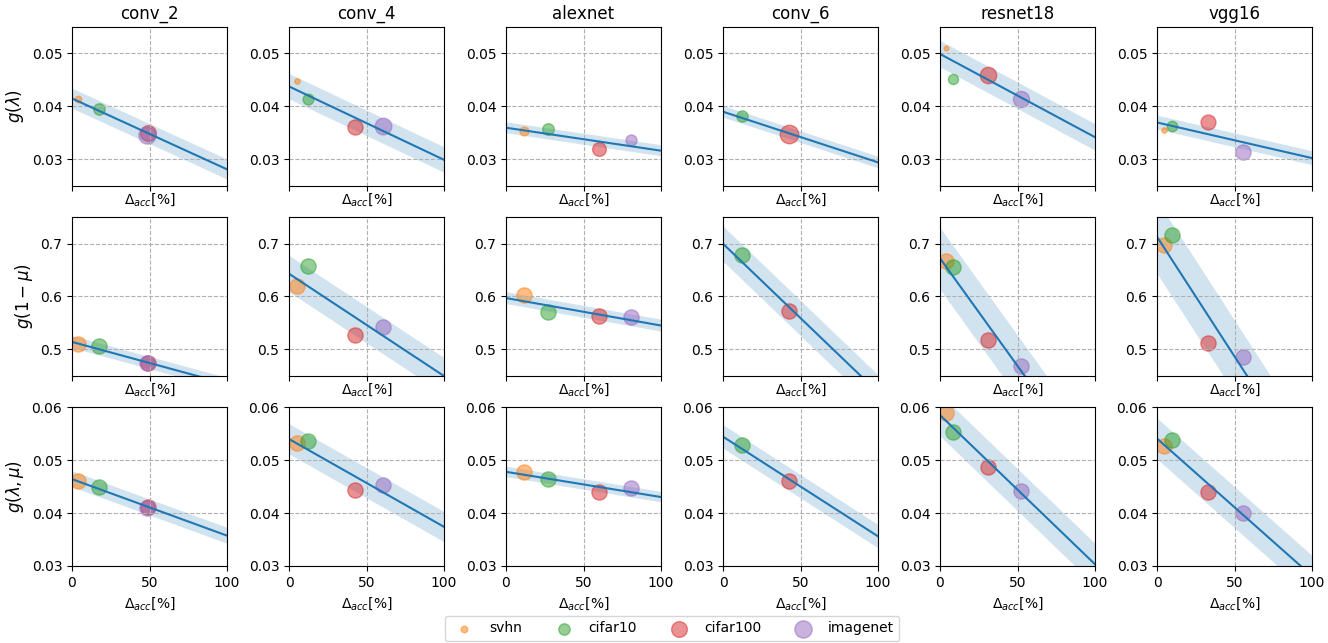}
    \caption{Topology summaries against performance (accuracy) gap for different models trained to recognize objects. Each disc represents mean (centre) and standard deviation (radius) on a particular dataset. Linear mapping and the corresponding standard deviation of the observed samples are marked.}
    \label{fig:results_object_recognition}
\end{figure*}

\subsection{Models}
We have chosen a wide range of architectures (i.e., topologies), including three standard and popular models \cite{krizhevsky2012imagenet, simonyan2014very, he2016deep} and a set of custom-designed ones. This provides diversity in depth, number of parameters, and topology. The custom DNNs are summarized in Table \ref{tab:networks}.

For semantic segmentation, we use a custom architecture called Fully Convolutional Network (FCN) capable of producing dense pixel predictions \cite{long2015fully}. It casts classical classifier networks \cite{he2016deep, simonyan2014very} into an encoder-decoder topology. The encoder can be any of the networks previously used.

\subsection{Training}

For all the datasets we have used, if a separate test set is provided, we attach it to the train data and perform a $k-fold$ cross validation on all available data. For object recognition $k=10$, and for the other two problems $k=5$. Each training is performed using a learning rate $\in \{0.1, 0.01, 0.001\}$ with random initialization. We also train each model on $100\%$, $50\%$ and $30\%$ of the available folds. This increases the generalization gap variance for a specific dataset. In the results presented below, we show all the trainings that achieved a performance metric above $50\%$. We skip extreme cases of generalization gaps either close to maximum ($95\%$).

For object recognition the input images are resized to $32\times32$ color images, unless explicitly stated. Also they are randomly cropped and randomly flipped. In the case of semantic segmentation all input images are $480 \times 480$ color images. No batch normalization (except for ResNet which follows the original design), dropout, or other regularization techniques are used during training. We train with a sufficient fixed number of epochs to guarantee saturation in both training and validation performance.

We use a standard stochastic gradient descent (SGD) optimizer for all training with momentum$=.9$ and learning rate and weight decay as indicated above. The learning rate is adaptive following a plateau criterion on the test performance, reducing to a quarter every time the validation performance metric does not vary outside a range for a fixed number of epochs.

\section{Results and Discussion}
\label{sec:results}

\begin{figure}
    \centering
    \includegraphics[width=\linewidth]{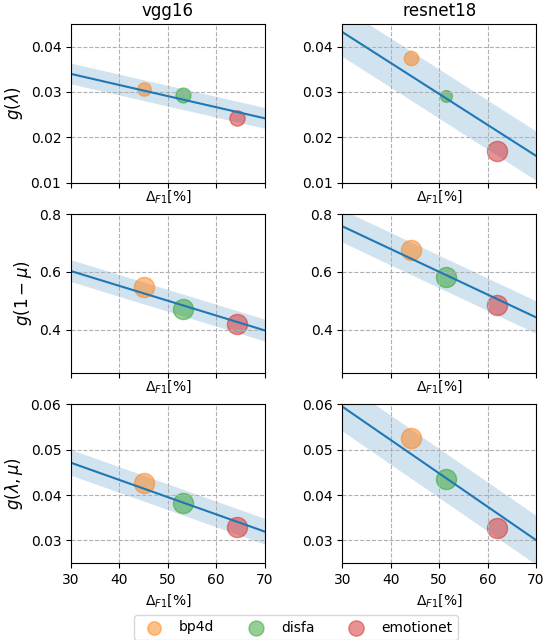}
    \caption{Topology summaries against performance (F1-score) gap for different models trained for AU recognition. Each disc represents mean (centre) and standard deviation (radius) for a 10-fold cross validation. Linear mapping and the corresponding standard deviation of the observed samples are marked.}
    \label{fig:results_au_recognition}
\end{figure}

\begin{table*}[]
    \centering
    \begin{tabular}{|c|c|c|c|c|c|c|c|}
    \hline
        Model & conv\_2 & conv\_4 & alexnet & conv\_6  & resnet18 & vgg16 & mean \\ \hline
        $g(\lambda)$ & $7.54\pm5.54$ & $10.12\pm6.93$ & $18.85\pm11.36$ & $6.70\pm4.25$ & $11.98\pm8.28$ & $16.01\pm8.00$ & $11.86\pm7.39$ \\ \hline
        $g(\mu)$ & $7.98\pm7.95$ & $12.69\pm5.59$ & $15.87 \pm10.25$ & $7.08\pm6.01$ & $6.79\pm4.65$ & $8.93\pm5.47$ & $9.89\pm6.65$ \\ \hline
        $g(\lambda, \mu)$ & $7.62\pm5.88$ & $9.45\pm6.88$ & $15.26\pm7.94$ & $6.57\pm5.80$ & $4.92\pm3.50$ & $6.93\pm5.10$ & $8.45\pm5.85$ \\ \hline
    \end{tabular}
    \caption{Evaluation of Alg. \ref{algo:alg1} for object recognition. Mean and standard deviation error (in \%) in estimating the test performance with leave-one-sample-out.}
    \label{tab:eval_or}
\end{table*}

\begin{table}[]
    \centering
    \small\setlength\tabcolsep{3pt}
    \begin{tabular}{|c|c|c|c|c|}\hline
        Model & conv2  & conv4 & vgg16 & resnet18 \\ \hline
        svhn & 10.94$\pm$4.99  & 10.73$\pm$2.60 & 8.01$\pm$2.31 & 11.09$\pm$4.18 \\ \hline
        cifar10 & 9.64$\pm$3.60 & 5.24$\pm$1.92 & 9.41$\pm$1.72 & 5.67$\pm$0.73 \\ \hline
        cifar100 & 4.79$\pm$6.20 & 22.46$\pm$6.87 & 10.93$\pm$1.46 & 6.71$\pm$1.51 \\ \hline
        imagenet & 8.33$\pm$5.48 & 21.84$\pm$7.35 & 13.49$\pm$7.14 & 9.49$\pm$4.35\\ \hline
    \end{tabular}%
    \caption{Evaluation of Alg. \ref{algo:alg1} for object recognition. Mean and standard deviation error (in \%) in estimating the test performance with leave-one-dataset-out. Each row indicates the dataset left out.}
    \label{tab:obj_rec_ldo}
\end{table}

\begin{table}[]
    \centering
    \begin{tabular}{|c|c|c|c|}
    \hline
        Model & resnet18 & vgg16 & mean \\ \hline
        $g(\lambda)$ & $6.96\pm4.18$ & $6.29\pm3.98$ & $6.62\pm4.08$ \\ \hline
        $g(\mu)$ & $4.02\pm3.51$ & $5.78\pm3.65$ & $3.83\pm3.58$ \\ \hline
        $g(\lambda, \mu)$ & $5.18\pm3.62$ & $5.87\pm3.62$ & $5.52\pm3.62$ \\ \hline
    \end{tabular}
    \caption{Evaluation of Alg. \ref{algo:alg1} for AU recognition. Mean and standard deviation error (in \%) in estimating the test performance with leave-one-sample-out.}
    \label{tab:eval_aur}
\end{table}

\begin{table}[]
   \centering
   \small
  \begin{tabular}{|c|c|c|}\hline
      Model & resnet18 & vgg16 \\ \hline
      bp4d & 3.82$\pm$2.80 & 6.04$\pm$4.17 \\ \hline
      disfa & 3.07$\pm$2.17 & 4.07$\pm$3.56 \\ \hline
      emotionet & 7.48$\pm$3.66 & 7.46$\pm$5.01 \\ \hline
  \end{tabular}
 \caption{Evaluation of Alg. \ref{algo:alg1} for object recognition. Mean and standard deviation error (in \%) in estimating the test performance with leave-one-dataset-out. Each row indicates the dataset left out.}
   \label{tab:au_rec_ldo}
\end{table}

\begin{figure}
    \centering
    \includegraphics[width=\linewidth]{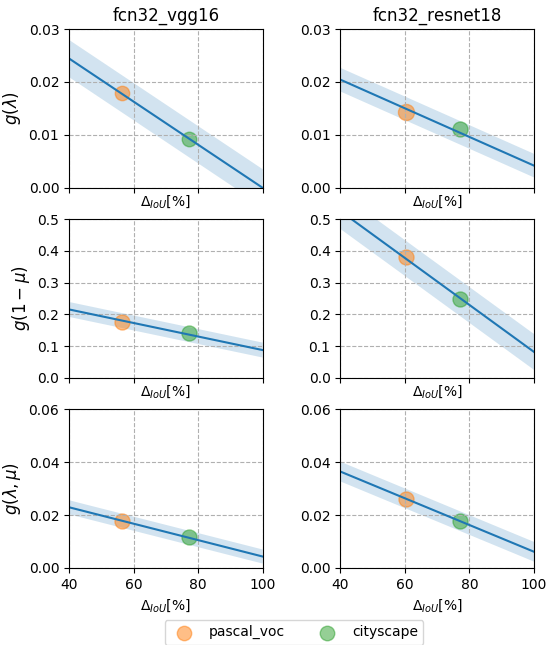}
    \caption{Topology summaries against performance (IoU) gap for different models trained for semantic segmentation.  Each disc represents mean (centre) and standard deviation (radius) on a particular dataset. Linear mapping and the corresponding standard deviation of the observed samples are marked.}
    \label{fig:results_ss}
\end{figure}

\begin{table}[]
    \centering
    \begin{tabular}{|c|c|c|c|}
    \hline
        Model & fcn32\_vgg16 & fcn32\_resnet18 & mean \\ \hline
         $g(\lambda)$ & $4.86\pm4.14$ & $6.01\pm4.34$ & $5.43\pm4.24$ \\ \hline
         $g(\mu)$ & $9.46\pm4.34$ & $4.68\pm3.82$ & $7.07\pm4.08$ \\ \hline
         $g(\lambda, \mu)$ & $5.60\pm2.86$ & $3.61\pm3.37$ & $4.60\pm3.11$ \\ \hline
    \end{tabular}
    \caption{Evaluation of Alg. \ref{algo:alg1} in semantic segmentation. Mean and standard deviation error (in \%) in estimating the test performance with leave-one-sample-out.}
    \label{tab:eval_ss}
\end{table}

\begin{table}[]
    \centering
    \resizebox{\linewidth}{!}{\begin{tabular}{|c|p{5cm}|c|}
    \hline
    Network & Convolutions & FC layers\\ \hline
    conv\_2 & $64_{3\times 3}$ $\rightarrow$ $64_{3\times 3}$ & 256, 256, $\#n\_classes$ \\ \hline
    conv\_4 & $64_{3\times 3}$ $\rightarrow$ $64_{3\times 3}$ $\rightarrow$ $64_{3\times 3}$ $\rightarrow$ $128_{3\times 3}$  & 256, 256, $\#n\_classes$ \\ \hline
    conv\_6 & $64_{3\times 3}$ $\rightarrow$ $64_{3\times 3}$  $\rightarrow$ $128_{3\times 3}$ $\rightarrow$ $128_{3\times 3}$ $\rightarrow$ $256_{3\times 3}$ $\rightarrow$ $256_{3\times 3}$ & 256, 256, $\#n\_classes$ \\ \hline
    \end{tabular}}
    \caption{An overview of the custom deep networks used in this paper. \emph{$conv\_x$} stands for convolutional networs with $x$ layers. We indicate each convolutional layer by its number of filters followed by their size in subscript and a pooling layer $P$ by its size in subscript. AlexNet \cite{krizhevsky2012imagenet}, VGG \cite{simonyan2014very}, and ResNet \cite{he2016deep} are standard architectures. Refer to original papers for details.}
    \label{tab:networks}
\end{table}

Topological summaries are strongly correlated with the performance gap. This holds true over different vision problems, datasets and networks.

\emph{Life} $\lambda$, the average persistence of cavities in the Vietoris-Rips filtration, is negatively correlated with the performance gap, with an average correlation of $-0.76$. This means that the more structured the functional metric space of the DNN (i.e., larger wholes it contains), the less it overfits.

\emph{Midlife} $\mu$ is positively correlated with the performance gap, with an average correlation of $0.82$. \emph{Midlife} is an indicator of the average distance $\epsilon$ about which cavities are formed. For DNNs that overfit less, cavities are formed at smaller $\epsilon$ which indicates that fewer connections in the metric spaces are needed to form them.

We show the plots of topological summaries against performance gap for object recognition, AU recognition and semantic segmentation in Figures \ref{fig:results_object_recognition}-\ref{fig:results_ss}, respectively. The linear mapping $g(\cdot)$ between each topological summary, life and midlife (Eqs. \ref{eq:life} \& \ref{eq:midlife}), and the performance gap are shown in the first and second rows of these figures, respectively. In all these figures rows represent DNN's results.

The results of each  dataset are indicated by a disc, where the centre specifies the mean  and the radius the standard deviation. We also mark the linear regression line $g(\cdot)$ and the corresponding standard deviation of the observed samples from it.

Finally, Tables \ref{tab:eval_or}, \ref{tab:eval_aur} \& \ref{tab:eval_ss} show the $Error$, namely the absolute value of the difference between the estimate given by Alg. \ref{algo:alg1} and that of a testing set, computed with Eq. (\ref{Eq:Error}) by leaving-one-sample-out. A different way of showing the same results can be found in Tables \ref{tab:obj_rec_ldo} and \ref{tab:au_rec_ldo} where mean and standard deviation of the same error is computed by leaving-one-dataset-out.

It is worth mentioning that our algorithm is general and can be applied to any DNN architecture. In Table \ref{tab:networks} we detail the structure of the networks that have been used in this paper. These networks range from simple (with only a few hundred of nodes, i.e., $conv_2$), to large (e.g., ResNet) nets with many thousands of nodes.

The strong correlations between basic properties of the functional graph of a DNN and fundamental learning properties like the performance gap also makes these networks more transparent. Not only do we propose an algorithm capable of computing the performance gap, but we show that this is linked to a simple law of the inner-workings of that networks. We consider this to be a contribution to make deep learning more explainable.

Based on these observations we have chosen to model the relationship between the performance gap and the topological summaries through a linear function. Figures \ref{fig:results_object_recognition}-\ref{fig:results_ss} show a simplified representation of the observed $(\lambda_i, \mu_i, \Delta_{\rho i})$ pairs and the regressed lines.

We need to mention that choosing a linear hypothesis for $g$ is by no means the only option. Obviously, using a non-linear regressor for $g(\cdot)$ in Alg. \ref{algo:alg1} leads to even more accurate predictions of the testing error. However, this improvement comes at the cost of being less flexible when studying less common networks/topologies -- overfitting.

Table \ref{tab:eval_or}-\ref{tab:eval_ss} further evaluate the algorithm proposed.

Crucially, an average error between $8.45\%$ and $4.60\%$ is obtained across computer vision problems, which is as accurate as computing a testing error with a labelled dataset $\mathcal{Z}$.

\section{Conclusions}

We have derived, to our knowledge, the first algorithm to compute the testing classification accuracy of any DNN-based system in computer vision, without the need for the collection of any testing data.

The main advantages of the proposed evaluation method versus the classical use of a testing dataset are:
\begin{enumerate}
    \item there is no need for a sequestered dataset to be maintained and updated by a third-party,
    \item there is no need to run costly cross-validation analyses,
    \item we can modify our DNN without the concern of overfitting to the testing data (because it does not exist), and,
    \item we can use all the available data for training the system.
\end{enumerate}

We have provided extensive evaluations of the proposed approach on three classical computer vision problems and shown the efficacy of the derived algorithm.

As a final note, we would like to point out the obvious. When deriving computer vision systems, practitioners would generally want to use all the testing tools at their disposal. The one presented in this paper is one of them, but we should not be limited by it. Where we  have access to a sequestered database, we should take advantage of it. In combination, multiple testing approaches should generally lead to better designs.


{\small {\bf Acknowledgments.}
NIH grants R01-DC-014498 and R01-EY-020834, Human Frontier Science Program RGP0036/2016, TIN2016-74946-P (MINECO/FEDER, UE), CERCA (Generalitat de Catalunya) and ICREA (ICREA Academia). CC and AMM defined main ideas and derived algorithms. CC, with SE, and MM ran experiments. CC and AMM wrote the paper. }


{\small
\bibliographystyle{ieee}
\bibliography{egbib}

\begin{thebibliography}{10}\itemsep=-1pt

\bibitem{arora2018stronger}
S.~Arora, R.~Ge, B.~Neyshabur, and Y.~Zhang.
\newblock Stronger generalization bounds for deep nets via a compression
  approach.
\newblock {\em arXiv preprint arXiv:1802.05296}, 2018.

\bibitem{barrett2019emotional}
L.~F. Barrett, R.~Adolphs, S.~Marsella, A.~M. Martinez, and S.~D. Pollak.
\newblock Emotional expressions reconsidered: Challenges to inferring emotion
  from human facial movements.
\newblock {\em Psychological Science in the Public Interest}, 20(1):1--68,
  2019.

\bibitem{bartlett2017spectrally}
P.~L. Bartlett, D.~J. Foster, and M.~J. Telgarsky.
\newblock Spectrally-normalized margin bounds for neural networks.
\newblock In {\em Advances in Neural Information Processing Systems}, pages
  6240--6249, 2017.

\bibitem{benitez2017emotionet}
C.~F. Benitez-Quiroz, R.~Srinivasan, Q.~Feng, Y.~Wang, and A.~M. Martinez.
\newblock Emotionet challenge: Recognition of facial expressions of emotion in
  the wild.
\newblock {\em arXiv preprint arXiv:1703.01210}, 2017.

\bibitem{bergomi2019towards}
M.~G. Bergomi, P.~Frosini, D.~Giorgi, and N.~Quercioli.
\newblock Towards a topological--geometrical theory of group equivariant
  non-expansive operators for data analysis and machine learning.
\newblock {\em Nature Machine Intelligence}, 1(9):423--433, 2019.

\bibitem{chazal2009gromov}
F.~Chazal, D.~Cohen-Steiner, L.~J. Guibas, F.~M{\'e}moli, and S.~Y. Oudot.
\newblock Gromov-hausdorff stable signatures for shapes using persistence.
\newblock In {\em Computer Graphics Forum}, volume~28, pages 1393--1403. Wiley
  Online Library, 2009.

\bibitem{cordts2016Cityscapes}
M.~Cordts, M.~Omran, S.~Ramos, T.~Rehfeld, M.~Enzweiler, R.~Benenson,
  U.~Franke, S.~Roth, and B.~Schiele.
\newblock The cityscapes dataset for semantic urban scene understanding.
\newblock In {\em Proc. of the IEEE Conference on Computer Vision and Pattern
  Recognition (CVPR)}, 2016.

\bibitem{corneanu2019does}
C.~A. Corneanu, M.~Madadi, S.~Escalera, and A.~M. Martinez.
\newblock What does it mean to learn in deep networks? and, how does one detect
  adversarial attacks?
\newblock In {\em Proceedings of the IEEE Conference on Computer Vision and
  Pattern Recognition}, pages 4757--4766, 2019.

\bibitem{deng2009imagenet}
J.~Deng, W.~Dong, R.~Socher, L.-J. Li, K.~Li, and L.~Fei-Fei.
\newblock Imagenet: A large-scale hierarchical image database.
\newblock In {\em Computer Vision and Pattern Recognition, 2009. CVPR 2009.
  IEEE Conference on}, pages 248--255. Ieee, 2009.

\bibitem{everingham2015pascal}
M.~Everingham, S.~A. Eslami, L.~Van~Gool, C.~K. Williams, J.~Winn, and
  A.~Zisserman.
\newblock The pascal visual object classes challenge: A retrospective.
\newblock {\em International journal of computer vision}, 111(1):98--136, 2015.

\bibitem{pascal-voc-2012}
M.~Everingham, L.~Van~Gool, C.~K.~I. Williams, J.~Winn, and A.~Zisserman.
\newblock The {PASCAL} {V}isual {O}bject {C}lasses {C}hallenge 2012 {(VOC2012)}
  {R}esults.
\newblock
  http://www.pascal-network.org/challenges/VOC/voc2012/workshop/index.html.

\bibitem{fabian2016emotionet}
C.~Fabian Benitez-Quiroz, R.~Srinivasan, and A.~M. Martinez.
\newblock Emotionet: An accurate, real-time algorithm for the automatic
  annotation of a million facial expressions in the wild.
\newblock In {\em Proceedings of the IEEE Conference on Computer Vision and
  Pattern Recognition}, pages 5562--5570, 2016.

\bibitem{hatcher2002algebraic}
A.~Hatcher.
\newblock {\em Algebraic Topology}.
\newblock Cambridge University Press, 2002.

\bibitem{he2016deep}
K.~He, X.~Zhang, S.~Ren, and J.~Sun.
\newblock Deep residual learning for image recognition.
\newblock In {\em Proceedings of the IEEE conference on computer vision and
  pattern recognition}, pages 770--778, 2016.

\bibitem{jiang2018predicting}
Y.~Jiang, D.~Krishnan, H.~Mobahi, and S.~Bengio.
\newblock Predicting the generalization gap in deep networks with margin
  distributions.
\newblock {\em arXiv preprint arXiv:1810.00113}, 2018.

\bibitem{krizhevsky2009learning}
A.~Krizhevsky and G.~Hinton.
\newblock Learning multiple layers of features from tiny images.
\newblock Technical report, Citeseer, 2009.

\bibitem{krizhevsky2012imagenet}
A.~Krizhevsky, I.~Sutskever, and G.~E. Hinton.
\newblock Imagenet classification with deep convolutional neural networks.
\newblock In {\em Advances in neural information processing systems}, pages
  1097--1105, 2012.

\bibitem{lecun2015deep}
Y.~LeCun, Y.~Bengio, and G.~Hinton.
\newblock Deep learning.
\newblock {\em nature}, 521(7553):436, 2015.

\bibitem{long2015fully}
J.~Long, E.~Shelhamer, and T.~Darrell.
\newblock Fully convolutional networks for semantic segmentation.
\newblock In {\em Proceedings of the IEEE conference on computer vision and
  pattern recognition}, pages 3431--3440, 2015.

\bibitem{mavadati2013disfa}
S.~M. Mavadati, M.~H. Mahoor, K.~Bartlett, P.~Trinh, and J.~F. Cohn.
\newblock Disfa: A spontaneous facial action intensity database.
\newblock {\em IEEE Transactions on Affective Computing}, 4(2):151--160, 2013.

\bibitem{netzer2011reading}
Y.~Netzer, T.~Wang, A.~Coates, A.~Bissacco, B.~Wu, and A.~Y. Ng.
\newblock Reading digits in natural images with unsupervised feature learning.
\newblock {\em Journal}, 2011.

\bibitem{neyshabur2017exploring}
B.~Neyshabur, S.~Bhojanapalli, D.~McAllester, and N.~Srebro.
\newblock Exploring generalization in deep learning.
\newblock In {\em Advances in Neural Information Processing Systems}, pages
  5947--5956, 2017.

\bibitem{rieck2018neural}
B.~Rieck, M.~Togninalli, C.~Bock, M.~Moor, M.~Horn, T.~Gumbsch, and
  K.~Borgwardt.
\newblock Neural persistence: A complexity measure for deep neural networks
  using algebraic topology.
\newblock {\em arXiv preprint arXiv:1812.09764}, 2018.

\bibitem{simonyan2014very}
K.~Simonyan and A.~Zisserman.
\newblock Very deep convolutional networks for large-scale image recognition.
\newblock {\em ICLR}, 2015.

\bibitem{vietoris1927hoheren}
L.~Vietoris.
\newblock {\"U}ber den h{\"o}heren zusammenhang kompakter r{\"a}ume und eine
  klasse von zusammenhangstreuen abbildungen.
\newblock {\em Mathematische Annalen}, 97(1):454--472, 1927.

\bibitem{vinyals2015show}
O.~Vinyals, A.~Toshev, S.~Bengio, and D.~Erhan.
\newblock Show and tell: A neural image caption generator.
\newblock In {\em Proceedings of the IEEE conference on computer vision and
  pattern recognition}, pages 3156--3164, 2015.

\bibitem{zador2019critique}
A.~M. Zador.
\newblock A critique of pure learning and what artificial neural networks can
  learn from animal brains.
\newblock {\em Nature communications}, 10(1):1--7, 2019.

\bibitem{zhang2014bp4d}
X.~Zhang, L.~Yin, J.~F. Cohn, S.~Canavan, M.~Reale, A.~Horowitz, P.~Liu, and
  J.~M. Girard.
\newblock Bp4d-spontaneous: a high-resolution spontaneous 3d dynamic facial
  expression database.
\newblock {\em Image and Vision Computing}, 32(10):692--706, 2014.

\bibitem{zomorodian2005computing}
A.~Zomorodian and G.~Carlsson.
\newblock Computing persistent homology.
\newblock {\em Discrete \& Computational Geometry}, 33(2):249--274, 2005.

\end{thebibliography}
}

\end{document}